\documentclass{aircc}

\usepackage{amssymb}
\usepackage{amsmath}
\usepackage{graphicx}
\usepackage[export]{adjustbox}
\usepackage{wrapfig}

\usepackage{url}

\newtheorem{definition}{Definition}
\newtheorem{theorem}{Theorem}

\newtheorem{lemma}{Lemma}

\begin{document}


\title{Fuzzy Logic in Narrow Sense with Hedges}


%
%
\author{Van-Hung Le}
%

\affiliation{Faculty of Information Technology\\Hanoi University of Mining and Geology, Vietnam}

%
%
\email{levanhung@humg.edu.vn}

\maketitle

\begin{abstract}
Classical logic has a serious limitation in that it cannot cope with the issues of vagueness and uncertainty into which fall most modes of human reasoning. In order to provide a foundation for human knowledge representation and reasoning in the presence of vagueness, imprecision, and uncertainty, fuzzy logic should have the ability to deal with linguistic hedges, which play a very important role in the modification of fuzzy predicates. In this paper, we extend fuzzy logic in narrow sense with graded syntax, introduced by Nov\'{a}k et al., with many hedge connectives. In one case, each hedge does not have any dual one. In the other case, each hedge can have its own dual one. The resulting logics are shown to also have the Pavelka-style completeness.
\end{abstract}

\section{Introduction}
Extending logical systems of mathematical fuzzy logic (MFL) with hedges is axiomatized by H\'{a}jek \cite{Hajek01}, Vychodil \cite{Vychodil06}, Esteva \emph{et al.} \cite{EGN13}, among others.
Hedges are called \emph{truth-stressing} or \emph{truth-depressing} if they, respectively, strengthen or weaken the meaning of the applied proposition. 
Intuitively, on a chain of truth values, the truth function of a truth-depressing (resp., truth-stressing) hedge (connective) is a superdiagonal (resp., subdiagonal) non-decreasing function preserving 0 and 1.
In \cite{Hajek01,Vychodil06,EGN13}, logical systems of MFL are extended by a truth-stressing hedge and/or a truth-depressing one.  

Nevertheless, in the real world, humans often use many hedges, e.g., \emph{very}, \emph{highly}, \emph{rather}, and \emph{slightly}, simultaneously to express different levels of emphasis. Furthermore, a hedge may or may not have a dual one, e.g., \emph{slightly} (resp., \emph{rather}) can be seen as a dual hedge of \emph{very} (resp., \emph{highly}).
Therefore, in \cite{LeLT14,LeT15}, Le \emph{et al.} propose two axiomatizations for propositional logical systems of MFL with many hedges. 
In the axiomatization in \cite{LeT15}, each hedge does not have any dual one whereas in the axiomatization in \cite{LeLT14}, each hedge can have its own dual one.
In \cite{LeT15,LeL15}, logical systems with many hedges for representing and reasoning with linguistically-expressed human knowledge are also proposed. 

Fuzzy logic in narrow sense with graded syntax (FLn) is introduced by Nov\'{a}k \emph{et al.} in \cite{NPM00}. In FLn, both syntax and semantics are evaluated by degrees. The graded approach to syntax can be seen as an elegant and natural generalization of classical logic for inference under vagueness since it allows one to explicitly represent and reason with partial truth, i.e., proving partially true conclusions from partially true premises, and it enjoys the Pavelka-style completeness. 
In this paper, we extend FLn with many hedges in order to provide a foundation for human knowledge representation and reasoning in the presence of vagueness since linguistic hedges are very often used by humans and play a very important role in the modification of fuzzy predicates. FLn is extended in two cases: (i) each hedge does not have a dual one, and (ii) each hedge can have its own dual one. 
We show that the resulting logics also have the Pavelka-style completeness.

The remainder of the paper is organized as follows. Section 2 gives an overview of notions and results of FLn. Section 3 presents two extensions of FLn with many hedges. In one case, each hedge does not have any dual one, and in the other, each hedge can have its own dual one. The resulting logics are also shown to have the Pavelka-style completeness. Section 4 concludes the paper.

\section{Fuzzy Logic in Narrow Sense}
FLn \cite{NPM00} is \emph{truth functional}. A compound formula is built from its constituents using a logical connective. The truth value of a compound formula is a function of the truth values of its constituents. The function is called the \emph{truth function} of the connective.

The set of truth values forms a residuated lattice, and more precisely, a \L ukasiewicz
algebra (or MV-algebra) $\mathcal{L}=\langle L, \vee, \wedge, \otimes, \Rightarrow, 0, 1 \rangle$, where $L=[0,1]$, in which:

(i) $\langle L, \vee, \wedge, 0, 1 \rangle$ is a lattice with the ordering $\leq$ defined using the operations $\vee$ (supremum), $\wedge$ (infimum) as usual, and $0, 1$ are its least and the greatest elements, respectively;

(ii) The operations $\otimes$ and $\Rightarrow$ are \L ukasiewicz conjunction and implication defined by $a\otimes b = 0 \vee (a+b-1)$ and $a\Rightarrow b = 1 \wedge (1-a+b)$, respectively. Thus, $a\leq b$ iff $a\Rightarrow b=1$.
They satisfy the \emph{residuation property} \cite{Ha98}: 
$a\otimes b \leq c \mbox{ iff } a\leq b \Rightarrow c$.

The language $J$ of FLn consists of: (\emph{a}) a countable set $Var$ of object variables $x, y,\dots $;
(\emph{b}) a finite or countable set of object constants $\mathbf{u_1}, \mathbf{u_2},\dots$;
(\emph{c}) a finite or countable set $Func$ of functional symbols $f, g,\dots$;
(\emph{d}) a nonempty finite or countable set $Pred$ of predicate symbols $P, Q,\dots$;
(\emph{e}) logical constants $\overline{a}$ for all $a\in L$;
(\emph{f}) implication connective $\rightarrow$;
(\emph{g}) general quantifier $\forall$; and
(\emph{h}) various types of brackets as auxiliary symbols.

\emph{Terms} are defined as follows:
(\emph{i}) a variable $x$ or constant $u$ is a (atomic) term;
(\emph{ii}) if $f$ be an \emph{n}-ary functional symbol and $t_l,\dots,t_n$ terms, then $f(t_l,\dots,t_n)$ is a term.

\emph{Formulae} are defined as follows:
(a) logical constants $\overline{a}$ are a formula; 
(b) if $P$ is an \emph{n}-ary predicate symbol, and $t_l,\dots,t_n$ are terms, then $P(t_l,\dots,t_n)$ is a formula;
(c) if $A, B$ are formulae, then $A\rightarrow B$ is a formula; and
(d) if $x$ is a variable, and $A$ is a formula, then $(\forall x)A$ is a formula.
Other connectives and formulae are defined as follows:
\begin{eqnarray*}
\neg A \equiv  A\rightarrow \overline{0} \; \mbox{(negation)}; \;\;
A\& B \equiv\neg (A\rightarrow \neg B) \;  \mbox{(\L ukasiewicz conjunction)};\\
A \vee B \equiv (B\rightarrow A)\rightarrow A \; \mbox{(disjunction)};\;\;
A\triangledown B \equiv \neg(\neg A\& \neg B) \; \mbox{(\L ukasiewicz disjunction)};\\
A\wedge B \equiv\neg((B\rightarrow A) \rightarrow \neg B) \;  \mbox{(conjunction)};\;\;
A \leftrightarrow B \equiv (A \rightarrow B) \wedge (B \rightarrow A) \;  \mbox{(equivalence)};\\
A^n \equiv\underbrace{A\& A \& \dots \& A}_{n-times} \;  \mbox{(n-fold conjunction)};\;\;
nA \equiv\underbrace{A\triangledown A \triangledown \dots \triangledown A}_{n-times} \;  \mbox{(n-fold disjunction)};\\
(\exists x) A \equiv \neg (\forall x) \neg A \;\;  \mbox{(existential quantifier)}.
\end{eqnarray*}
The set of all formulae of $J$ is denoted by $F_J$. 

An \emph{evaluated formula} is a pair $a/A$, where $A \in F_J$ and $a \in L$ is its syntactic evaluation.
\emph{Axioms} are sets of evaluated formulae. Since the evaluations can be interpreted as membership degrees in the fuzzy set, axioms can be seen as fuzzy sets of formulae. 

$A \Subset U$ denotes that $A$ is a fuzzy set on a universe $U$. The set of all fuzzy sets on $U$ is denoted by $\mathcal{F}(U) = \{A|A\Subset U\}$. 
$\mathcal{F}(U)$ also contains all ordinary subsets of $U$.

Since a fuzzy set can be considered as a function, if we have a function $V : F_J\Rightarrow L$ and a fuzzy set $W\Subset F_J$, then $V \leq W$ means the ordering of functions.

\begin{definition} \cite{NPM00}
An \emph{n}-ary \emph{inference rule} r in FLn is of the form:
\begin{equation}
r: \;\; \frac{a_1/A_1, \dots, a_n/A_n}{r^{evl}(a_1, \dots,a_n)/r^{syn}(A_1,\dots,A_n)}
\end{equation}
which means from $a_1/A_1, \dots, a_n/A_n$ infer $r^{evl}(a_1, \dots,a_n)/r^{syn}(A_1,\dots,A_n)$, where $r^{syn}$ is a partial \emph{n}-ary \emph{syntactic} operation on $F_J$, and $r^{evl}$ is an \emph{n}-ary lower semicontinuous \emph{evaluation} operation on $L$ (i.e., it preserves arbitrary suprema in all variables).
\end{definition}
A fuzzy set $V\Subset F_J$ is \emph{closed} w.r.t. $r$ if
$V(r^{syn}(A_1, \dots,A_n))\geq r^{evl}(V(A_1),\dots,V(A_n))$
holds for all formulae $A_1,\dots,A_n \in Dom(r^{syn})$.

The set $R$ of inference rules of FLn consists of the following:
\begin{eqnarray*}
&&\mbox{Modus ponens:}\;\; r_{MP}:\;\;\;\frac{a/A, b/A\rightarrow B}{a\otimes b/B}; \;\;\;\;\;
\mbox{Generalization:}\;\; r_{G}:\;\;\; \frac{a/A}{a/(\forall x)A} \\
&&\mbox{Logical constant introduction:}\;\; r_{LC}:\;\;\; \frac{a/A}{a\Rightarrow a/\overline{a} \rightarrow A} 
\end{eqnarray*}
Note that the evaluation operation $r_{LC}^{evl}(x)$ of $r_{LC}$ is $a \Rightarrow x$. 

An \emph{proof} of $A\in F_J$ from a fuzzy set $X\Subset F_J$ is a finite sequence of evaluated formulae $w:= a_0/A_0, a_1/A_1, \dots, a_n/A_n$,
whose each member is either a member of $X$, i.e., $a_i/A_i:=\;\;\; X(A_i)/A_i$, or follows from some preceding members of the sequence using an inference rule of FLn, and the last member is $a_n/A_n:=a/A$. The evaluation $a$ is called the \emph{value} of the proof $w$, denoted $Val(w)$. A proof $w$ of a formula $A$ can be denoted $w_A$. 

A \emph{graded consequence operation} is a closure operation $\mathcal{C} : \mathcal{F}(F_J)\Rightarrow\mathcal{F}(F_J)$ assigning to a fuzzy set $X\Subset F_J$ a fuzzy set $\mathcal{C}(X)\Subset F_J$ and fulfilling $\mathcal{C}(X)=\mathcal{C}(\mathcal{C}(X))$. 
\begin{definition} \cite{NPM00}
Let $R$ be the set of inference rules. The fuzzy set of \emph{syntactic consequences}
of a fuzzy set $X \Subset F_J$ is the following membership function, for all $A\in F_J$:
\begin{equation}
\mathcal{C}^{syn}(X)(A)=\bigwedge\{V(A)|V\Subset F_J, X\leq V \mbox{ and } V \mbox{ is closed w.r.t. all } r \in R\}
\end{equation}
\end{definition}
\begin{theorem} \cite{NPM00} \label{ct47}
Let $X \Subset F_J$. Then, $\mathcal{C}^{syn}(X)(A)=\bigvee\{Val(w)|w \mbox{ is a proof of } A \mbox{ from } X\}$.
\end{theorem}

A \emph{structure} for the language $J$ of FLn is $\mathcal{D}=\langle D, (P_D)_{P\in Pred}, (f_D)_{f\in Func}, u_1, u_2, \dots\rangle$, where $D$ is a non-empty domain (set); 
$(P_D)\Subset D$ assigns to each \emph{n}-ary predicate symbol $P\in Pred$ an \emph{n}-ary fuzzy relation $P_D$ on $D$; 
$(f_D)$ assigns to each \emph{n}-ary functional symbol $f\in Func$ an \emph{n}-ary function $f_D$ on $D$;
$u_1,u_2\dots \in D$ are designated elements which are assigned to each constant $\mathbf{u_1},\mathbf{u_2},\dots$ of the language $J$, respectively.

A \emph{truth valuation} of formulae in a structure $\mathcal{D}$ is a function (also denoted by) $\mathcal{D}: F_J \rightarrow \mathcal{L}$ defined by means of \emph{interpretation}.
Let $\mathcal{D}$ be a structure for the language $J$. The language $J(\mathcal{D})$ is obtained from $J$ by adding new constants being names for all elements from $D$, i.e., $J(\mathcal{D}) = J \cup \{\mathbf{d}| d \in D\}$. 

Let $A(x)$ be a formula and $t$ a term. $A_x[t]$ denotes a formula obtained from $A$ by replacing all free occurrences of the variable $x$ with $t$.

\emph{Interpretations} of closed terms and formulae are defined as follows:

(i) \emph{Interpretation} of closed terms: $\mathcal{D}(\mathbf{u_i})=u_i$ if $\mathbf{u_i}\in J$ and $u_i \in D$; $\mathcal{D}(\mathbf{d})=d$ if $d\in D$; $\mathcal{D}(f(t_1,\dots,t_n))=f_D(\mathcal{D}(t_1),\dots,\mathcal{D}(t_n))$.

(ii) \emph{Interpretation} of closed formulae (where $t_1,\dots,t_n$ are closed terms): 
$\mathcal{D}(\overline{a}) = a$ for all $a \in L$;
$\mathcal{D}(P(t_1,\dots,t_n)) = P_D (\mathcal{D}(t_1),\dots,\mathcal{D}(t_n))$;
$\mathcal{D}((\forall x)A) = \bigwedge\{\mathcal{D}(A_x[\mathbf{d}])| d \in D\}$.

(iii) \emph{Interpretation} of the derived connectives:
\begin{eqnarray*}
&\mathcal{D}(\neg A) = \neg \mathcal{D}(A) &
\mathcal{D}(A \wedge B) = \mathcal{D}(A) \wedge \mathcal{D}(B)\\
&\mathcal{D}(A \& B) = \mathcal{D}(A) \otimes \mathcal{D}(B)&
\mathcal{D}(A \vee B) = \mathcal{D}(A) \vee \mathcal{D}(B)\\
&\mathcal{D}(A \triangledown B) = \mathcal{D}(A) \oplus \mathcal{D}(B)&
\mathcal{D}(A \leftrightarrow B) = \mathcal{D}(A) \Leftrightarrow \mathcal{D}(B)\\
&\mathcal{D}((\exists x)A) = \bigvee\{\mathcal{D}(A_x[\mathbf{d}])| d \in D\}
\end{eqnarray*} 
where $\neg$ and $\oplus$ are \L ukasiewicz negation and disjunction defined by  $\neg a = 1 - a$ and $a\oplus b = 1 \wedge (a+b)$, respectively.

If $A(x_1,\dots,x_n)$ is not a closed formula, an \emph{evaluation}
 of its free variables $x_1,\dots,x_n$ is first defined such that $e(x_1) = d_1,\dots,
e(x_n) = d_n$. Then, the interpretation of $A$ is the interpretation of
$A_{x_1,\dots,x_n}[\mathbf{d}_1,\dots,\mathbf{d}_n]$, which is a closed formula.

\begin{definition} \cite{NPM00}
Let $X \Subset F_J$ be a fuzzy set of formulae. Then the fuzzy set of its \emph{semantic} consequences is the following membership function:
\begin{equation}
\mathcal{C}^{sem}(X)(A)=\bigwedge\{\mathcal{D}(A)|\mbox{ for all structure } \mathcal{D}, X\leq \mathcal{D}\}.
\end{equation}
\end{definition}
A formula $A$ is an \emph{a-tautology} (tautology in the degree $a$) if
$a=\mathcal{C}^{sem}(\emptyset)(A)$, and it is denoted by $\models_a A$. If $a=1$, it is simply written by $\models A$, and $A$ is called a \emph{tautology}.

The following lemma gives simple rules how to verify tautologies in FLn.
\begin{lemma} \cite{NPM00} \label{lem44}
Let $A, B$ be formulae in the language $J$.

(a) $\models A \rightarrow B$ iff $\mathcal{D}(A)\leq \mathcal{D}(B)$ holds for every structure $\mathcal{D}$ for the language $J$.

(b) $\models A \leftrightarrow B$ iff $\mathcal{D}(A)= \mathcal{D}(B)$ holds for every structure $\mathcal{D}$ for the language $J$.
\end{lemma}

\begin{definition} \cite{NPM00} Let (R1) $A\rightarrow (B \rightarrow A)$; 
(R2) $(A\rightarrow B) \rightarrow ((B\rightarrow C)\rightarrow (A\rightarrow C))$;

(R3) $(\neg B \rightarrow \neg A)\rightarrow (A\rightarrow B)$;
(R4) $((A\rightarrow B)\rightarrow B)\rightarrow ((B\rightarrow A)\rightarrow A)$;

(B1) $(\overline{a}\rightarrow \overline{b})\leftrightarrow \overline{(a\Rightarrow b)}$, where $\overline{(a\Rightarrow b)}$ denotes the logical constant for the value $a\Rightarrow b$ when $a$ and $b$ are given. This is called \emph{book-keeping axiom};

(T1) $(\forall x)A\rightarrow A_x[t]$ for any substitutible term $t$. This is called \emph{substitution axiom};

(T2) $(\forall x)(A\rightarrow B)\rightarrow (A\rightarrow (\forall x)B)$ provided that $x$ is not free in $A$.
\end{definition}
Using Lemma \ref{lem44}, it can be verified that \emph{(R1)-(R4), (B1)} and \emph{(T1)-(T2)} are (1-)tautologies.

The fuzzy set $LAx$ of \emph{logical axioms} of FLn is as follows: 
$LAx(F)=1$ if $F$ is one of the forms \emph{(R1)-(R4), (B1)} and \emph{(T1)-(T2)}; $LAx(F)=a$ if $F=\overline{a}$; and $LAx(F)=0$ otherwise.

\begin{definition} \cite{NPM00}
A \emph{fuzzy theory} (or \emph{theory} for short) $T$ in the language $J$ of FLn is a triple $T=\langle LAx,SAx,R\rangle$,
where $LAx$ is the fuzzy set of logical axioms, $SAx \Subset F_J$ is a fuzzy set of
\emph{special axioms}, and $R$ is the set of inference rules.
\end{definition}
A theory can be viewed as a fuzzy set $T = \mathcal{C}^{syn}(LAx \cup
SAx) \Subset F_J$.

\begin{definition} \cite{NPM00}
Let $T$ be a theory and $A\in F_J$ a formula.

(i) If $\mathcal{C}^{syn}(LAx\cup SAx)(A)=a$, it is denoted by $T\vdash_a A$, and $A$ is said to be a \emph{theorem} or \emph{provable} in the degree $a$ in $T$. The value $a$ is called the \emph{provability degree} of $A$ over $T$.

(ii) If $\mathcal{C}^{sem}(LAx\cup SAx)(A)=a$, it is denoted by $T\vDash_a A$, and $A$ is said to be \emph{true} in the degree $a$ in $T$. The value $a$ is called the \emph{truth degree} of $A$ over $T$.

(iii) Let $\mathcal{D}$ be a truth valuation of formulae. Then, it is a \emph{model} of $T$, denoted $\mathcal{D}\vDash T$, if $SAx(A)\leq \mathcal{D}(A)$ holds for all formulae $A \in F_J$.
\end{definition}
Therefore, Theorem \ref{ct47} can be restated as follows:
\begin{equation}
T\vdash_a A \mbox{ iff } a=\bigvee\{Val(w)|w \mbox{ is a proof of } A \mbox{ from } LAx\cup SAx\} \end{equation}
Also, due to the assumption made on LAx, we have $LAx(A) \leq \mathcal{D}(A)$ holds for every truth valuation $\mathcal{D}$ of formulae. Thus,
\begin{equation}
T \vDash_a A \mbox{ iff } a = \bigwedge\{\mathcal{D}(A)|\mathcal{D} \vDash T\}. \label{ct413}
\end{equation}
\begin{definition} \cite{NPM00}
A theory T is \emph{contradictory} if there exists a formula $A$, and there are proofs $w_A$ and
$w_{\neg A}$ of $A$ and $\neg A$ from $T$, respectively, such that
$Val(w_A) \otimes Val(w_{\neg A})>0$. Otherwise, it is \emph{consistent}.
\end{definition}
\begin{theorem} \cite{NPM00}
A fuzzy theory $T$ is consistent iff it has a model.
\end{theorem}
\begin{theorem} [Completeness] \cite{NPM00} \label{comthr}
$T\vdash_a A \mbox{ iff } T \models_a A$ holds for every formula $A \in F_J$ and every consistent fuzzy theory $T$.
\end{theorem}
This means that the provability degree of $A$ in $T$ coincides with its truth degree over $T$. This is usually referred to as the \emph{Pavelka-style completeness} \cite{Pa79,Ha98}.
\section{Fuzzy Logic in Narrow Sense with Hedges}
In order to provide a foundation for a computational approach to human reasoning in the presence of vagueness, imprecision, and uncertainty, fuzzy logic should have the ability to deal with linguistic hedges, which play a very important role in the modification of fuzzy predicates. Therefore, in this section, we will extend FLn with hedges connectives in order to model human knowledge and reasoning. In addition to extending the language and the definition of formulae, new logical axioms characterizing properties of the new connectives are added. One of the most important properties should be preservation of the logical equivalence.

\begin{definition} \cite{NPM00} \label{deflf}
Let $\diamond: L^n \rightarrow L$ be an n-ary operation. It is called \emph{logically fitting} if it satisfies the following condition: there are natural numbers $k_1 > 0, \dots ,k_n > 0$
such that 
\begin{equation*}
(a_1 \Leftrightarrow b_1)^{k_1} \otimes \dots \otimes (a_n \Leftrightarrow b_n)^{k_n} \leq \diamond(a_1, \dots, a_n) \Leftrightarrow \diamond(b_1, \dots, b_n)
\end{equation*}
holds for all $a_1,\dots,a_n, b_1,\dots, b_n \in L$, and the power is taken w.r.t. the operation $\otimes$, e.g., 
\begin{equation*}
(a_1 \Leftrightarrow b_1)^{k_1} = \underbrace{(a_1 \Leftrightarrow b_1)\otimes \dots \otimes (a_1 \Leftrightarrow b_1)}_{k_1-times}
\end{equation*}
\end{definition}
Hence, logically fitting operations preserve the logical equivalence.
It can be proved that all the basic operations $\vee, \wedge, \otimes, \Rightarrow$ are logically fitting, and any composite operation obtained from logically fitting operations is also logically fitting.

A connective is \emph{logically fitting} if it is assigned a logically fitting truth function (operation).

\subsection{FLn with many hedges}
In this subsection, FLn is extended with many hedges, in which each hedge does not have any dual one. To ease the presentation, we let $s_0, d_0$ denote the identity connective, i.e., for all formula $A$, $A\equiv s_0 A \equiv d_0 A$, and their truth functions $s^{\bullet}_0$ and $d^{\bullet}_0$ are the identity.

\begin{definition} [FLn with many hedges]
On the syntactic aspect, FLn is extended as follows (where $p,q$ are positive integers):

(i) The language $J$ is extended into a language $J_h$ by a finite set $\mathcal{H} = \{s_1, \dots, s_p, d_1, \dots, d_q\}$ of additional unary connectives, where $s_i$'s are truth-stressing hedges, and $d_j$'s are truth-depressing ones.

(ii) The definition of formulae is extended by adding the following: If $A$ is a formula and $h\in \mathcal{H}$, $hA$ is a formula.

(iii) The fuzzy set $LAx$ of logical axioms is extended by the following axioms:
\begin{eqnarray}
1/(A\rightarrow B) \rightarrow (hA \rightarrow hB), \mbox{ for all } h\in \mathcal{H} \label{ndp}\\
1/s_i A \rightarrow s_{i-1} A, \mbox{ for } i=1,\dots,p \label{tsp}\\
1/s_p \overline{1} \label{p1p}\\
1/d_{j-1} A \rightarrow d_{j} A, \mbox{ for } j=1,\dots,q \label{tdp}\\
1/\neg d_{q} \overline{0} \label{p0p}
\end{eqnarray}
\end{definition}
Axiom (\ref{ndp}) states that if $A$ implies $B$, then \emph{very} (resp., \emph{slightly}) $A$ implies \emph{very} (resp., \emph{slightly}) $B$. Axiom (\ref{p1p}) (resp., (\ref{p0p})) says that the truth function $s_p^{\bullet}$ (resp., $d_q^{\bullet}$) preserves 1 (resp., 0).
Axiom (\ref{tsp}) (resp., (\ref{tdp})) expresses that $s_i$ (resp., $d_j$) modifies truth more than $s_{i-1}$ (resp., $d_{j-1}$), for $i = 2,\dots,p$ (resp., $j = 2,\dots,q$). 
For example, \emph{slightly} (resp., \emph{very}) modifies truth more than \emph{rather} (resp., \emph{highly}) since \emph{slightly true} $<$ \emph{rather true} $<$ \emph{true} (resp., \emph{true} $<$ \emph{highly true} $<$ \emph{very true}). Also, for instance, let $A=\mathit{young}(x), s_1=\mathit{highly}, s_2=\mathit{very}, d_1=\mathit{rather}, d_2 = \mathit{slightly}$, by (\ref{tsp}), we have $\mathit{very}\,  \mathit{young}(x)\rightarrow \mathit{highly} \, \mathit{young}(x)$ and $\mathit{highly}\, \mathit{young}(x)\rightarrow \mathit{young}(x)$. Moreover, by (\ref{tdp}), we have $\mathit{young}(x)\rightarrow \mathit{rather}\, \mathit{young}(x)$ and $\mathit{rather}\, \mathit{young}(x)\rightarrow \mathit{slightly}\, \mathit{young}(x)$. Therefore, we have:
\begin{equation*}
\mathit{very}\, \mathit{young}(x)\rightarrow \mathit{highly}\, \mathit{young}(x) \rightarrow \mathit{young}(x) \rightarrow \mathit{rather}\, \mathit{young}(x) \rightarrow \mathit{slightly}\, \mathit{young}(x)
\end{equation*}
This is also in accordance with fuzzy-set-based interpretations of hedges \cite{Zadeh72,HuynhHN02,CockK04}, in which  \emph{very} and \emph{highly} are called \emph{intensifying modifiers} while \emph{rather} and \emph{slightly} are called \emph{weakening modifiers}, and they satisfy the so-called \emph{semantic entailment} property: 
\begin{equation*}
x \mbox{ is } \mathit{very}\, A \Rightarrow x \mbox{ is } \mathit{highly}\, A \Rightarrow x \mbox{ is } A \Rightarrow x \mbox{ is } \mathit{rather}\, A \Rightarrow x \mbox{ is } \mathit{slightly}\, A \end{equation*}
where \emph{A} is a fuzzy predicate. Note that, according to \cite{Zadeh72}, if $A$ is represented by a fuzzy set with a membership function $\mu_A(x)$, the membership function of $very\, A$ can be $\mu_{very\, A}(x)=\mu^2_A(x)$. Since for all $x, 0\leq \mu_A(x)\leq 1$, we have for all $x, \mu_{very\, A}(x)\leq \mu_A(x)$. By fuzzy set inclusion, it is said that $very\, A$ is included by $A$, denoted $very\, A\subseteq A$. Since the degree of membership of $x$ to $A$ is regarded as the truth value of ``\emph{x} is \emph{A}", the truth value of ``\emph{x} is \emph{very A}" is less than or equal to that of ``\emph{x} is \emph{A}".

\begin{theorem}
For every hedge connective $h \in \mathcal{H}$, its truth function $h^{\bullet}$  is non-decreasing and preserves 0 and 1.
\end{theorem}
\emph{Proof.}
By (\ref{ndp}), for all structure $\mathcal{D}$ of the language $J_h$, we have $\mathcal{D}(A \rightarrow B) \leq \mathcal{D}(hA \rightarrow hB)$. Hence, $\mathcal{D}(A) \Rightarrow \mathcal{D}(B) \leq h^{\bullet}(\mathcal{D}(A)) \Rightarrow h^{\bullet}(\mathcal{D}(B))$. Let $a, b \in L$ and $a\leq b$. Since $\mathcal{D}(\overline{a}) \Rightarrow \mathcal{D}(\overline{b})=a\Rightarrow b=1$, we have $h^{\bullet}(\mathcal{D}(\overline{a})) \Rightarrow h^{\bullet}(\mathcal{D}(\overline{b}))=1$, i.e., $h^{\bullet}(a)\leq h^{\bullet}(b)$. Therefore, the truth function $h^{\bullet}$ of any hedge connective $h \in \mathcal{H}$ is non-decreasing.

By (\ref{p1p}), we have $s_p^{\bullet}(1)=1$. Using (\ref{tsp}) and taking $A=\overline{1}, i=p$, we have, for all structure $\mathcal{D}$ of the language $J_h$, $\mathcal{D}(s_p\overline{1}) \leq \mathcal{D}(s_{p-1}\overline{1})$, i.e, $1=s_p^{\bullet}(1) \leq s_{p-1}^{\bullet}(1)$. Hence, $s_{p-1}^{\bullet}(1)=1$. Similarly, we have $s_{i}^{\bullet}(1)=1$ for all $i=p-2,\dots, 1$. 
Also, using (\ref{tsp}) and taking $A=\overline{0}, i=1$, we have, for all structure $\mathcal{D}$ of the language $J_h$, $\mathcal{D}(s_1\overline{0}) \leq \mathcal{D}(s_{0}\overline{0})$, i.e, $s_1^{\bullet}(0) \leq s_{0}^{\bullet}(0)=0$. Hence, $s_{1}^{\bullet}(0)=0$. Similarly, we have $s_{i}^{\bullet}(0)=0$ for all $i=2,\dots, p$. Therefore, all $s_{i}^{\bullet}$ preserve 0 and 1.

By (\ref{p0p}), we have $d_q^{\bullet}(0)=0$. Using (\ref{tdp}) and taking $A=\overline{0}, j=q$, we have, for all structure $\mathcal{D}$ of the language $J_h$, $\mathcal{D}(d_{q-1}\overline{0}) \leq \mathcal{D}(d_{q}\overline{0})$, i.e, $d_{q-1}^{\bullet}(0) \leq d_{q}^{\bullet}(0)=0$. Hence, $d_{q-1}^{\bullet}(0)=0$. Similarly, we have $d_{j}^{\bullet}(0)=0$ for all $j=q-2,\dots, 1$.
Also, using (\ref{tdp}) and taking $A=\overline{1}, j=1$, we have, for all structure $\mathcal{D}$ of the language $J_h$, $\mathcal{D}(d_{0}\overline{1}) \leq \mathcal{D}(d_{1}\overline{1})$, i.e, $1=d_{0}^{\bullet}(1) \leq d_{1}^{\bullet}(1)$. Hence, $d_{1}^{\bullet}(1)=1$. Similarly, we have $d_{j}^{\bullet}(1)=1$ for all $j=2,\dots, q$. Therefore, all $d_{j}^{\bullet}$ preserve 0 and 1. $\Box$

\begin{theorem}
For every truth-stresser $s_i \in \mathcal{H}, i = \overline{1,p}$, its truth function $s_i^{\bullet}$  is subdiagonal.
\end{theorem}
\emph{Proof.}
For any $a\in L$, using (\ref{tsp}) and taking $A=\overline{a}, i=1$, we have, for all structure $\mathcal{D}$ of the language $J_h$, $s_1^{\bullet}(a)=\mathcal{D}(s_1\overline{a}) \leq \mathcal{D}(s_{0}\overline{a})=s_0^{\bullet}(a)=a$. Thus, $s_1^{\bullet}(a) \leq a$ for all $a\in L$, i.e., $s_1^{\bullet}$ is subdiagonal. Again, using (\ref{tsp}) and taking $A=\overline{a}, i=2$, we have $s_2^{\bullet}(a)=\mathcal{D}(s_2\overline{a}) \leq \mathcal{D}(s_{1}\overline{a})=s_1^{\bullet}(a)\leq a$. Similarly, $s_i^{\bullet}$ is subdiagonal for all $i=3,\dots, p$. $\Box$

\begin{theorem}
For every truth-depresser $d_j \in \mathcal{H}, j = \overline{1,q}$, its truth function $d_j^{\bullet}$  is superdiagonal.
\end{theorem}
\emph{Proof.}
For any $a\in L$, using (\ref{tdp}) and taking $A=\overline{a}, j=1$, we have, for all structure $\mathcal{D}$ of the language $J_h$, $a=d_0^{\bullet}(a)=\mathcal{D}(d_0\overline{a}) \leq \mathcal{D}(d_{1}\overline{a})=d_1^{\bullet}(a)$. Thus, $d_1^{\bullet}(a) \geq a$ for all $a\in L$, i.e., $d_1^{\bullet}$ is superdiagonal. Again, using (\ref{tdp}) and taking $A=\overline{a}, j=2$, we have 
$a\leq d_1^{\bullet}(a)=\mathcal{D}(d_1\overline{a}) \leq \mathcal{D}(d_{2}\overline{a})=d_2^{\bullet}(a)$. Similarly, $d_j^{\bullet}$ is superdiagonal for all $j=3,\dots, q$. $\Box$

The following theorem shows that the additional hedge connectives are logically fitting.
\begin{theorem}
For every hedge connective $h \in \mathcal{H}$, its truth function $h^{\bullet}$  is logically fitting.
\end{theorem}
\emph{Proof.}
Given $a, b \in L$, using (\ref{ndp}) and taking $A=\overline{a}, B=\overline{b}$, we have, for all structure $\mathcal{D}$ of the language $J_h$, $\mathcal{D}(\overline{a}\rightarrow \overline{b}) \leq \mathcal{D}(h\overline{a}\rightarrow h\overline{b})$. Hence, $a\Rightarrow b\leq h^{\bullet}(a)\Rightarrow h^{\bullet}(b)$. Again, using (\ref{ndp}) and taking $A=\overline{b}, B=\overline{a}$, we have $b\Rightarrow a\leq h^{\bullet}(b)\Rightarrow h^{\bullet}(a)$. Therefore, $a\Leftrightarrow b = 
(a\Rightarrow b) \wedge (a\Leftarrow b)
\leq (h^{\bullet}(a)\Rightarrow h^{\bullet}(b)) \wedge (h^{\bullet}(a)\Leftarrow h^{\bullet}(b)) = h^{\bullet}(a)\Leftrightarrow h^{\bullet}(b)$, i.e., $h^{\bullet}$ is logically fitting according to Definition \ref{deflf}.  $\Box$

We can also say that all the hedge connectives $h\in \mathcal{H}$ are logically fitting.

Since it is shown in \cite{NPM00} that introducing logically fitting operations
have no influence on the algebraic proof of the completeness theorem of FLn (Theorem \ref{comthr}), we have the following completeness theorem for FLn with many hedges.
\begin{theorem}
Let $T$ be a consistent fuzzy theory in the extended language $J_h$. Then 
\begin{equation*}
T\vdash_a A \mbox{ iff } T \models_a A
\end{equation*}
holds for every formula $A \in F_{J_h}$.
\end{theorem}
That means FLn with many hedges also has the Pavelka-style completeness.
\subsection{FLn with many dual hedges}
It can be observed that each hedge can have a dual one, e.g., \emph{slightly} and \emph{rather}
can be seen as a dual hedge of \emph{very} and \emph{highly}, respectively. Thus, there might
be axioms expressing dual relations of hedges in addition to axioms expressing their comparative truth modification strength.

In this subsection, FLn is extended with many dual hedges, in which each hedge has its own dual one.
\begin{definition} [FLn with many dual hedges]
On the syntactic aspect, FLn is extended as follows (where $n$ is a positive integer):

(i) The language $J$ is extended into a language $J_{dh}$ by a finite set $\mathcal{H} = \{s_1, \dots, s_n, d_1, \dots, d_n\}$ of additional unary connectives, where $s_i$'s are truth-stressing hedges, and $d_i$'s are truth-depressing ones.

(ii) The definition of formulae is extended by adding the following: If $A$ is a formula and $h\in \mathcal{H}$, $hA$ is a formula.

(iii) The fuzzy set $LAx$ of logical axioms is extended by the following axioms:
\begin{eqnarray}
1/(A\rightarrow B) \rightarrow (hA \rightarrow hB), \mbox{ for all } h\in \mathcal{H} \label{ndp1}\\
1/s_i A \rightarrow s_{i-1} A, \mbox{ for } i=1,\dots,n \label{tsp1}\\
1/s_n \overline{1} \label{p1p1}\\
1/d_{i-1} A \rightarrow d_{i} A, \mbox{ for } i=1,\dots,n \label{tdp1}\\
1/d_{i}A \rightarrow \neg s_i \neg A \label{dp1}
\end{eqnarray}
\end{definition}
The main differences between the fuzzy set $LAx$ of FLn with many dual hedges and that of FLn with many hedges are that we have the axiom (\ref{dp1}) expressing the duality between hedges $s_i$ and $d_i$, and we do not need an axiom similar to (\ref{p0p}).

Concerning Axiom (\ref{dp1}), for instance, let $A=young(x), s_i = very, d_i = slightly$, then it means ``\emph{slightly young} implies not \emph{very old}''. Using (\ref{dp1}) and taking $A=\overline{a}$ for any $a\in L$, we have $\mathcal{D}(d_i\overline{a}) \leq \mathcal{D}(\neg s_i \neg \overline{a})$ for all structure $\mathcal{D}$ of the language $J_{dh}$. Thus, $d_i^{\bullet}(a) \leq 1-s_i^{\bullet}(1-a)$, then $s_i^{\bullet}(1-a)\leq 1-d_i^{\bullet}(a)$. Let $b=1-a=\neg a$, hence $a=1-b=\neg b$. We have $s_i^{\bullet}(b)\leq 1-d_i^{\bullet}(\neg b)$. Let $B$ be a formula whose truth valuation is $b$, i.e., $\mathcal{D}(B)=b$. We have $\mathcal{D}(s_iB)=s_i^{\bullet}(b)\leq 1-d_i^{\bullet}(\neg b)=\mathcal{D}(\neg d_i\neg B)$. Therefore, we have a tautology $1/s_{i}B \rightarrow \neg d_i \neg B$. This means, for instance, ``\emph{very old} implies not \emph{slightly young}'' as well.

Since axioms (\ref{tsp1}) and (\ref{p1p1}) are respectively similar to (\ref{tsp}) and (\ref{p1p}), we also have that all $s^{\bullet}_i, i=\overline{1,n},$ preserve 1. Using (\ref{dp1}) and taking $A=\overline{0}$, we have $\mathcal{D}(d_i\overline{0}) \leq \mathcal{D}(\neg s_i \neg \overline{0})$ for all structure $\mathcal{D}$ of the language $J_{dh}$. Hence, $d_i^{\bullet}(0) \leq 1-s_i^{\bullet}(1)=0$. Therefore, $d_i^{\bullet}(0)=0$ for all $i=\overline{1,n}$. That means we do not need an axiom similar to (\ref{p0p}).

Now, given a hedge function $s_n^{\bullet}$, which is non-decreasing, subdiagonal,
and preserves 0 and 1, the boundaries for the other hedge functions are one by one determined as
follows. 
Using (\ref{tsp1}) and taking $A=\overline{a}$ for any $a\in L$ and $i=n$, we have $s_n^{\bullet}(a)\leq s_{n-1}^{\bullet}(a)$. Thus, the lower boundary of $s_{n-1}^{\bullet}(a)$ is $s_n^{\bullet}(a)$, and its upper boundary is $a$. Similarly, we have $s_{n-1}^{\bullet}(a)\leq s_{n-2}^{\bullet}(a)\leq a$, and finally, $s_{2}^{\bullet}(a)\leq s_{1}^{\bullet}(a)\leq a$. 
Then, by (\ref{dp1}), we have $a\leq d_i^{\bullet}(a)\leq \neg s_i^{\bullet}(\neg a)$ for all $i=\overline{1,n}$, thus, $a$ and $\neg s_{1}^{\bullet}(\neg a)$  are the lower boundary and upper boundary of $d_1^{\bullet}(a)$, respectively (note that since $s_{1}^{\bullet}(1-a)\leq 1-a$, we have $\neg s_{1}^{\bullet}(\neg a) = 1-s_{1}^{\bullet}(1-a)\geq 1-(1-a) = a$). Using (\ref{tdp1}) and taking $A=\overline{a}, i=2$, we have $d_1^{\bullet}(a) \leq d_2^{\bullet}(a)$, thus, $d_1^{\bullet}(a)$ and $\neg s_{2}^{\bullet}(\neg a)$ are the lower boundary and upper boundary of $d_2^{\bullet}(a)$, respectively. Similarly, we have $d_{i-1}^{\bullet}(a)$ and $\neg s_{i}^{\bullet}(\neg a)$ are the lower boundary and upper boundary of $d_i^{\bullet}(a)$, respectively, for all $i=\overline{3,n}$. In summary, the boundaries are shown in Table \ref{Tab1}.

\begin{table} 
\centering
\caption{Boundaries of hedge functions} \label{Tab1}
\begin{tabular}{|c|c|c|} \hline
Hedge function& Lower boundary& Upper boundary\\ \hline
$s^{\bullet}_{n-1}(x)$ & $s^{\bullet}_n(x)$ & $x$\\ \hline
\dots&&\\ \hline
$s^{\bullet}_1(x)$ & $s^{\bullet}_2(x)$ & $x$\\ \hline
$d^{\bullet}_1(x)$ & $x$ & $-s^{\bullet}_1(-x)$\\ \hline
$d^{\bullet}_2(x)$ & $d^{\bullet}_1(x)$ & $-s^{\bullet}_2(-x)$\\ \hline
\dots&&\\ \hline
$d^{\bullet}_n(x)$ & $d^{\bullet}_{n-1}(x)$ & $-s^{\bullet}_n(-x)$\\
\hline\end{tabular}
\end{table}
We also have the following completeness theorem for FLn with many dual hedges.
\begin{theorem}
Let $T$ be a consistent fuzzy theory in the extended language $J_{dh}$ . Then 
\begin{equation*}
T\vdash_a A \mbox{ iff } T \models_a A
\end{equation*}
holds for every formula $A \in F_{J_{dh}}$.
\end{theorem}
It can be seen that in a case when we want to extend the above logic with
one truth-stressing (resp., truth-depressing) hedge without a dual one, we only need to add axioms expressing its relations to the existing truth-stressing (resp., truth-depressing) hedges
according to their comparative truth modification strength.
\section{Conclusion}
In this paper, we extend fuzzy logic in narrow sense with many hedge connectives in order to provide a foundation for human knowledge representation and reasoning.  
In fuzzy logic in narrow sense, both syntax and semantics are evaluated by degrees in [0,1]. The graded approach to syntax can be seen as an elegant and natural generalization of classical logic for inference under vagueness since it allows one to explicitly represent and reason with partial truth, i.e., proving partially true conclusions from partially true premises, and it enjoys the Pavelka-style completeness. 
FLn is extended in two cases: (i) each hedge does not have a dual one, and (ii) each hedge can have its own dual one.
In addition to extending the language and the definition of formulae, new logical axioms characterizing properties of the hedge connectives are added. 
The truth function of a truth-depressing (resp., truth-stressing) hedge connective is a superdiagonal (resp., subdiagonal) non-decreasing function preserving 0 and 1.
The resulting logics are shown to have the Pavelka-style completeness.

\section*{Acknowledgments}
Funding from HUMG under grant number T16-02 is acknowledged.


\vspace{5mm}
\textbf{AUTHOR}
\vspace{5mm}

\begin{wrapfigure}{r}{3cm}
\centering
\includegraphics[width=2.1cm,height=2.8cm]{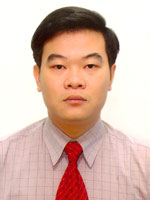} 
\end{wrapfigure}
\textbf{Van-Hung Le} received a B.Eng. degree in Information Technology from Hanoi University of Science and Technology (formerly, Hanoi University of Technology) in 1995, an M.Sc. degree in Information Technology from Vietnam National University, Hanoi in 2001, and a PhD degree in Computer Science from La Trobe University, Australia in 2010. 
He is a lecturer in the Faculty of Information Technology at Hanoi University of Mining and Geology.
His research interests include Fuzzy Logic, Logic Programming, Soft Computing, and Computational Intelligence. 
\end{document}